\documentclass[sigconf]{acmart}
\usepackage{multirow}
\usepackage{float}
\usepackage{stfloats}
\copyrightyear{2023}
\acmYear{2023}
\setcopyright{acmlicensed}
\acmConference[MM '23] {Proceedings of the 31st ACM International Conference on Multimedia}{October 29--November 3, 2023}{Ottawa, ON, Canada.}
\acmBooktitle{Proceedings of the 31st ACM International Conference on Multimedia (MM '23), October 29--November 3, 2023, Ottawa, ON, Canada}
\acmPrice{15.00}
\acmISBN{979-8-4007-0108-5/23/10}
\acmDOI{10.1145/3581783.3612042}
\begin{document}

\title{Multi-Frame Self-Supervised Depth Estimation with Multi-Scale Feature Fusion in Dynamic Scenes}



\author{Jiquan Zhong}
\email{zhongjiquan@stu.xmu.edu.cn}
\affiliation{%
  \institution{Department of Automation, \\Xiamen University,}
  \city{Xiamen}
  \country{China}
}

\author{Xiaolin Huang}
\email{xiaolinhuang@sjtu.edu.cn}
\affiliation{%
  \institution{Department of Automation, \\Shanghai Jiao Tong University,\\Key Laboratory of
System Control and Information Processing, \\Ministry of Education of China, }
  \city{Shanghai}
  \country{China}
}

\author{Xiao Yu}
\authornote{Corresponding author.}
\email{xiaoyu@xmu.edu.cn}
\affiliation{%
  \institution{Department of Automation, \\Xiamen University,\\ \!Key Laboratory of Multimedia Trusted Perception and Efficient Computing, \\Ministry of Education of China, }
  \city{Xiamen}
  \country{China}
  }



\begin{abstract}
  Monocular depth estimation is a fundamental task in computer vision and multimedia. The self-supervised learning pipeline makes it possible to train the monocular depth network with no need of depth labels. 
  In this paper, a multi-frame depth model with multi-scale feature fusion is proposed for strengthening texture features and spatial-temporal features, which improves the robustness of depth estimation between frames with large camera ego-motion. A novel dynamic object detecting method with geometry explainability is proposed. The detected dynamic objects are excluded during training, which guarantees the static environment assumption and relieves the accuracy degradation problem of the multi-frame depth estimation. Robust knowledge distillation with a consistent teacher network and reliability guarantee is proposed, which improves the multi-frame depth estimation without an increase in computation complexity during the test. The experiments show that our proposed methods achieve great performance improvement on the multi-frame depth estimation.
\end{abstract}

\ccsdesc[500]{Computing methodologies~Scene understanding}
\ccsdesc[300]{Computing methodologies~Vision for robotics}
\ccsdesc[300]{Computing methodologies~3D imaging}

\keywords{monocular depth estimation; self-supervised learning; dynamic detection; neural networks}



\maketitle

\section{Introduction}
\label{sec:intro}

\begin{figure}[t]
    \centering
    \includegraphics[width=0.95\linewidth]{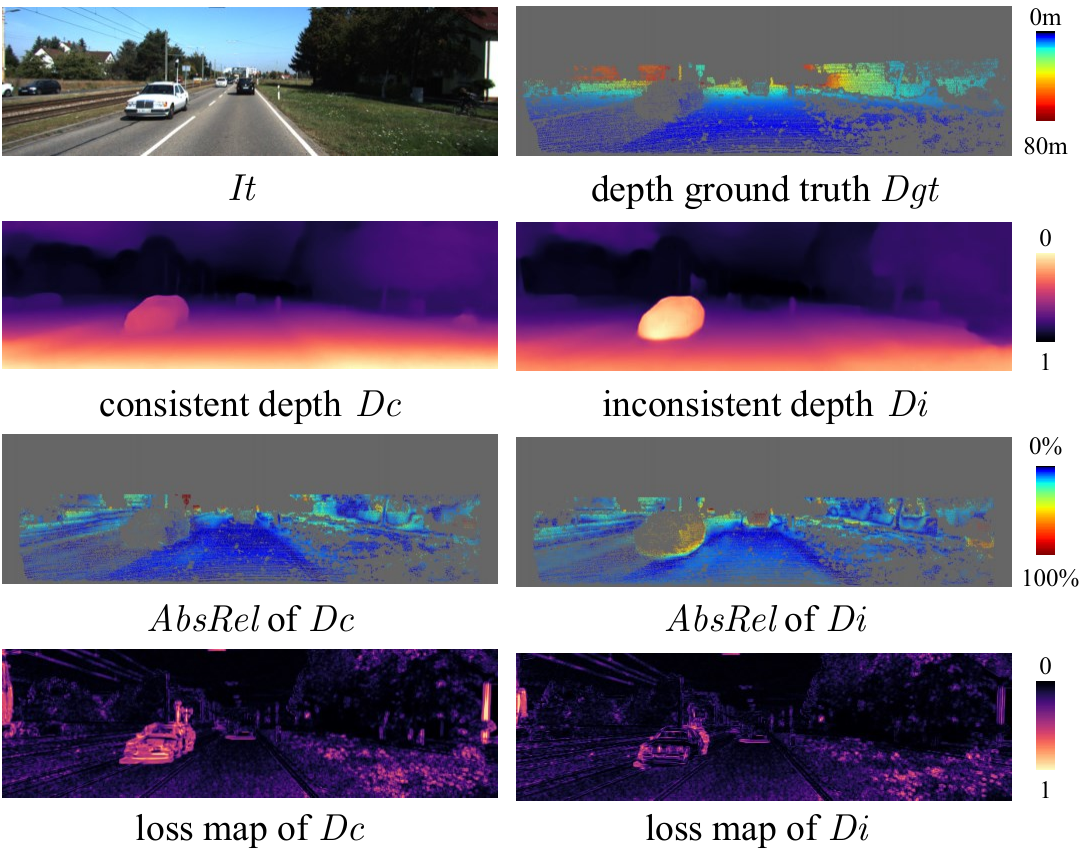}
    \vspace{-4pt}
    \caption{The inconsistent depth estimation learned with plain self-supervised loss. The under/over-estimation in dynamic regions leads to lower training loss but higher evaluation absolute relative error. $I_t$: input image; $D_{gt}$: depth ground truth from the velodyne; $D_c$: normalized consistent depth estimation; $D_i$: normalized inconsistent depth estimation, $AbsRel$: absolute relative error of the depth. (The depth maps are shown as normalized depth since the monocular depth estimation lacks scale information.)}
    \label{fig:main}
\end{figure}

Environmental depth perception~\cite{zhou2022learning,xu2022multi,uhrig2017sparsity} is significant for autonomous mobile agents, such as autonomous vehicles and mobile robots. Similar to the ability of humans to perceive depth from visual information, obtaining depth estimation from RGB images is a more appealing way than directly from the specialized hardware such as Lidar sensors. Recent self-supervised methods enable learning monocular depth from arbitrary unlabelled image sequences~\cite{garg2016unsupervised}. While the conventional single-frame depth estimation method takes a single image as input and predicts the corresponding depth per pixel~\cite{vijayanarasimhan2017sfm}, the multi-frame method achieves better performance by considering temporal adjacent images as input~\cite{watson2021temporal}. The training procedure for both the single-frame and multi-frame methods is based on the assumption of a static environment for the self-supervised loss. However, there are many dynamic objects in real-world scenarios, which leads to degradation of the accuracy of the learned monocular depth model~\cite{yin2018geonet}. As shown in Figure~\ref{fig:main},
training a monocular depth model with the plain self-supervised loss will cause the under/over-estimation problems in dynamic regions, since the inconsistent depth estimation will lead to lower self-supervised loss  in the dynamic regions compared to the consistent depth estimation. 
The embedded spatial-temporal feature in the multi-frame depth network brings performance improvements in static regions, but it also brings heavier depth inconsistency problems in the dynamic scenes, since the spatial-temporal feature contains more information about the dynamic regions than the single-frame texture feature. To relieve this problem, disentangling dynamic objects~\cite{feng2022disentangling} is proposed. However, this disentangling method requires additional semantic information and considers static vehicles and pedestrians as dynamic objects, which leads to unnecessary computational resources. Other works tend to use transformer~\cite{guizilini2022multi} and attention modules~\cite{ruhkamp2021attention} to improve feature matching, which improves the depth estimation but increases resource consumption largely.

\begin{figure*}
    \centering
     \vspace{-10pt}
    \includegraphics[width=1.0\linewidth]{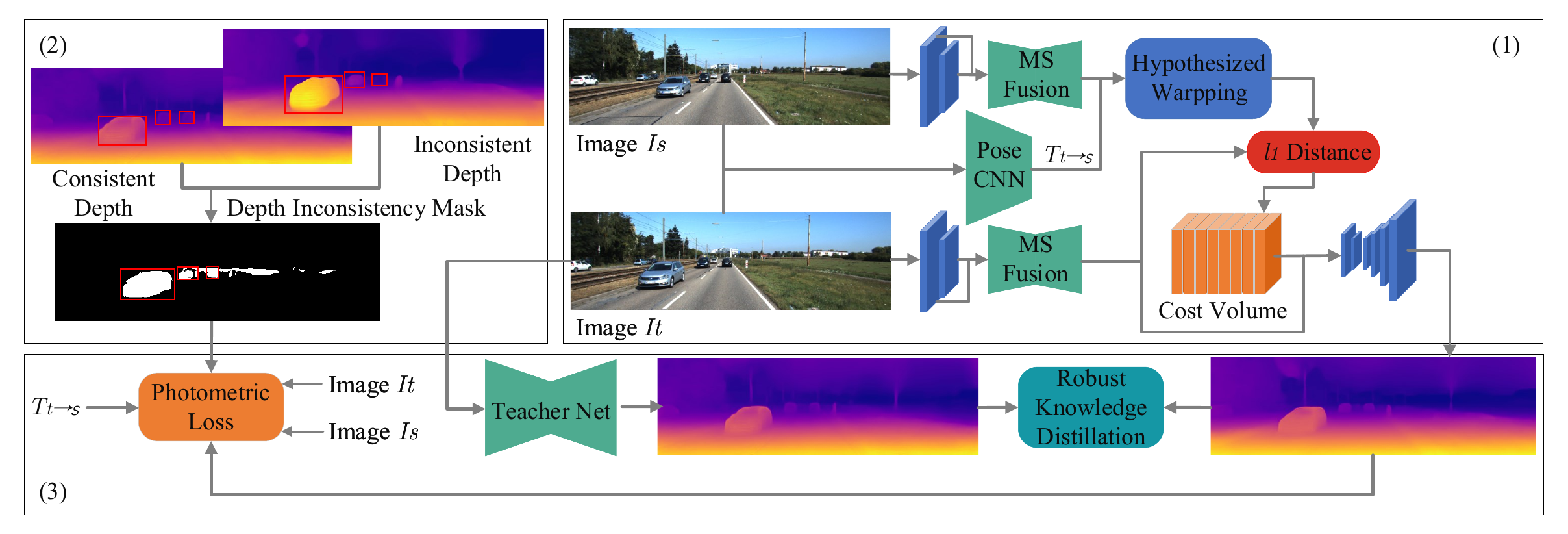}
    \vspace{-20pt}
    \caption{An overview of our proposed self-supervised multi-frame depth learning framework, including (1) multi-frame depth network with multi-scale feature fusion, (2) depth inconsistency mask generation, and (3) self-supervised training with robust knowledge distillation.
    }
    \vspace{-10pt}
    \label{fig:mainwork}
\end{figure*}

In this paper, a self-supervised monocular depth learning framework is proposed for multi-frame depth estimation in dynamic scenes.
As shown in Figure~\ref{fig:mainwork}, the multi-frame depth network with multi-scale feature aggregation and a pose network are joint learning with depth inconsistency masks filtering out the dynamic regions and robust knowledge distillation for regularization.
Overall, the contribution of this paper is stated as follows:
\begin{itemize}
	\item \emph{Multi-frame depth network with feature aggregation} 
    We propose a multi-frame depth network that strengthens both the spatial-temporal feature and the texture feature. A multi-scale feature fusion block is proposed to enhance the characterization of multi-scale features, which improves the robustness of depth estimation during larger camera ego-motion.

	\item \emph{Depth inconsistency mask} The dynamic regions break the static environment assumption and mislead both the depth prediction and the pose prediction. We leverage the inconsistent performance of the multi-frame depth estimation and propose a depth inconsistency mask to filter out the dynamic regions during training. Our depth inconsistency mask is off-the-shelf and can be used in other frameworks.

	\item \emph{Robust knowledge distillation} To improve the multi-frame depth prediction in and dynamic regions, a single-frame depth network is used for pseudo supervision. We propose an approach which uses a consistent teacher network and filters out unreliable pseudo-depth labels to achieve robust knowledge distillation.
	
\end{itemize}

\section{Related work}
\label{sec:relatework}

\subsection{Self-supervised monocular depth learning}

Self-supervised monocular depth learning~\cite{zhou2017unsupervised} aims to use unlabeled monocular videos to train a monocular depth network. To obtain depth label loss during training, a photo-metric loss is applied between the target frames and reconstructed frames. However, the photo-metric loss usually failed at the occlusion, moving objects, texture-less, and illumination variance regions, causing the photo-metric loss to be based on the assumption of a static environment~\cite{zhou2017unsupervised}. Many methods are proposed to fix the artifacts of the raw photo-metric loss, for instance, minimal photo-metric loss between multiple source frames~\cite{godard2019digging}, leveraging segmentation information during training~\cite{lee2021attentive}, geometry-based occlusion masking~\cite{bian2019unsupervised}, illumination alignment~\cite{yang2020d3vo}, and objects motion detection~\cite{li2021unsupervised}. The camera intrinsic is usually needed during training, but the requirement can be loosened~\cite{gordon2019depth}.

\subsection{Multi-frame monocular depth estimation}

Similar to the traditional SLAM methods~\cite{newcombe2011dtam}, self-supervised monocular depth methods can use multiple frames to refine the depth estimation during testing. Different from the SLAM methods that directly refine depth estimation and ego-motion estimation, the test-time-refinement method fine-tunes the depth network and pose network using the same photo-metric losses as the training. However, in need of iterations of forward and backward computation, it costs much more time for the test-time-refinement method~\cite{chen2019self} to obtain final depth estimation.

Another method to leverage the multi-frame information is to use the recurrent network architecture~\cite{patil2020don,wang2019recurrent}. However, these recurrent monocular depth network lacks the geometry formation embeddings. The spatial-temporal features are implicitly learned by the neural network which limits the representation capacity of the recurrent depth networks.

The more efficient methods to leverage the multiple frames during training and testing are multi-view-stereo (MVS) based monocular depth networks. ManyDepth~\cite{watson2021temporal} and DepthFormer~\cite{guizilini2022multi} use ego-motion estimation and hypothesized depth bins to perform feature matching between adjacent frames. The aggregated spatial-temporal features together with the texture features are fed into the decoder for depth estimation. Different from directly using the $l1$ cost volume as the spatial-temporal features as ManyDepth~\cite{watson2021temporal}, DepthFormer~\cite{guizilini2022multi} adopts the grouped self-attention architecture to obtain the cost volume and achieve more robust spatial-temporal aggregation. TC-depth~\cite{ruhkamp2021attention} proposes a method using spatial attention and temporal attention to aggregate the spatial-temporal information and achieves scale invariant depth estimation with geometry consistency losses. 

\subsection{Self-supervised monocular depth learning in dynamic scenes}

Real-world scenarios are full of dynamic objects such as pedestrians, vehicles, bicycles, etc. The violation from the dynamic objects to the assumption of static objects in self-supervised depth learning needs to be relieved. CC~\cite{ranjan2019competitive} and Zhou et al.~\cite{zhou2017unsupervised} propose a method using a network to predict irregular pixels and assign low weights on the re-projection loss. Lee et al.~\cite{lee2021attentive} use the semantic box to detect dynamic objects. DynamicDepth~\cite{feng2022disentangling} proposes to use single-frame depth as prior depth, to relieve the motion of the semantic objects for a multi-frame depth network. Li et al.~\cite{li2021unsupervised} and RM-Depth~\cite{hui2022rm} propose to use a network to predict the motion of the dynamic objects and disentangle the static scenes and the dynamic objects on the re-projection loss. DiPE~\cite{jiang2020dipe} proposes an outlier mask to filter out the dynamic objects. However, the semantic-based methods require semantic labels or pre-trained semantic segmentation networks. The object's motion prediction-based methods need to train an extra network coupled with the re-projection loss.  

\section{Method}

In this section, we first introduce the main framework of self-supervised monocular depth learning (Sec.~\ref{sec:self}). We then introduce three innovations that improve the monocular depth estimation: (1) Multi-frame depth model with multi-scale feature fusion (Sec.~\ref{sec:revisit}), (2) depth inconsistency masks (Sec.~\ref{sec:overfitting}), and (3) robust knowledge distillation (Sec.~\ref{sec:robust}).

\subsection{Self-supervised monocular depth learning}
\label{sec:self}

The joint self-supervised learning of the monocular depth and the camera ego-motion takes monocular videos to train a depth network and an ego-motion network. Our main pipeline is mainly based on~\cite{godard2019digging}.
Pose network estimates the camera ego-motion between two adjacent frames: $N_P(I_t, I_s) \rightarrow T_{t \rightarrow s}$. Multi-frame depth network estimates depth from adjacent frames and camera ego-motion: $N_{MD}(I_t, I_s, T_{t \rightarrow s}) \rightarrow \hat{D}_t$. The joint learning of the depth network and pose network is the procedure of mainly minimizing the reconstruction photo-metric loss constructed by $l1$ norm and SSIM~\cite{wang2004image}:
\begin{equation}
    \mathcal{PE}(I_t,I_s) = \frac{\alpha}{2}(1-SSIM(I_t, I_{s \rightarrow t})) + (1-\alpha) ||I_t- I_{s \rightarrow t}||_1,
    \label{loss:proj}
\end{equation}
with $\alpha=0.85$. In Eq.~\ref{loss:proj}, $I_{s \rightarrow t}$ is the reconstructed target frame $I_t$ from source frames $I_s$:
\begin{equation}
    I_{s \rightarrow t} = I_s < proj(\hat{D}_t, T_{t \rightarrow s}, K)>,
    \label{eqn:proj}
\end{equation}
where $K$ is the camera intrinsic, $<>$ is the bi-linear sampling operator, and $proj(\cdot)$ is the camera re-projection function~\cite{zhou2017unsupervised} getting the corresponding coordinates determined by $\hat{D}_t$ and $T_{t \rightarrow s}$. In order to filter out the unreliable occlusion regions, the per-pixel minimum reconstruction loss between multiple source frames is adopted~\cite{godard2019digging}:
\begin{equation}
    L_{ph} = \min_s \mathcal{PE}(I_t,I_s),
\end{equation}
where $s\in \{t-1,t+1\}$ is used for training. The smoothness loss is adopted for regularization~\cite{zhou2017unsupervised}:
\begin{equation}
    L_{sm} = |\partial_x {\hat{d}_t}|e^{-\partial_x I_t} + |\partial_y {\hat{d}_t}|e^{-\partial_y I_t},
\end{equation}
where $\hat{d_t}$ is mean-normalized inverse depth.

\subsection{Multi-frame depth model with multi-scale feature fusion}
\label{sec:revisit}

\subsubsection{Main structure}
Our multi-frame model uses the ResNet-18~\cite{he2016deep} as the encoder. The first two layers of ResNet-18 are used as the texture feature encoder $N_f$ and the last three layers are used as depth encoder $N_e$. The depth decoder $N_d$ is the same as~\cite{godard2019digging}. The multi-frame depth network leverages the multi-view-stereo method to embed the spatial-temporal features. The target frame $I_t$ and source frame $I_s$ are first encoded to be pyramidal features $F_1$ and $F_2$ by the feature encoder $N_f$ separately. The features $F_2$ at $1/4$ raw resolution are used as texture features $F_t$ and $F_s$ to balance the computation complexity and performance. Similar to  Eq.~\ref{eqn:proj}, given the pose estimation $ T_{t \rightarrow s}$, the source features from view $V_{t-1}$ are re-projected to the target view $V_t$ as:
\begin{equation}
    F^i_{s \rightarrow t} = F_s < proj(d^i, T_{t \rightarrow s}, K)>,
    \label{eqn:featproj}
\end{equation}
where $d^i$ is the hypothesized depth which is obtained by discretizing the sampling in linear space, ranging from $d_{\min}$ to $d_{\max}$, into $D$ bins. Each $d^i$ is given by:
\begin{equation}
    d^i = d_{\min} + \frac{i(d_{\max}-d_{\min})}{D-1},\ i=0,1,\cdots,D-1.
\end{equation}
The $l1$ norm of $F_t - F^i_{s \rightarrow t}$ in the feature channel for each $d^i$ are aggregated to obtain the cost volume $C_{s\rightarrow t}\in R^{D\times H/4 \times W/4}$:
\begin{equation}
    C^i_{s\rightarrow t} = \frac{1}{n} \sum_{j=1}^{n} || F_t - F^{i,j}_{s \rightarrow t} ||_0,
\end{equation}
where $i=0,1,\cdots,D-1$ and $n$ is the number of feature channels. The texture features $F_t$ and cost volume $C_{s\rightarrow t}$ are then fed into the depth encoder $N_e$ and decoder $N_d$ to generate multi-frame depth. As the photo-metric loss, the lower cost volume indicating the corresponding hypothesized depth is more likely to be the depth ground truth. This also means that the cost volume tends to be unreliable in inconsistent regions such as occlusion and dynamic regions.

\subsubsection{Multi-scale feature fusion} 
\label{sec:msff}
\begin{figure}[t]
    \centering
    \includegraphics[width=1.0\linewidth]{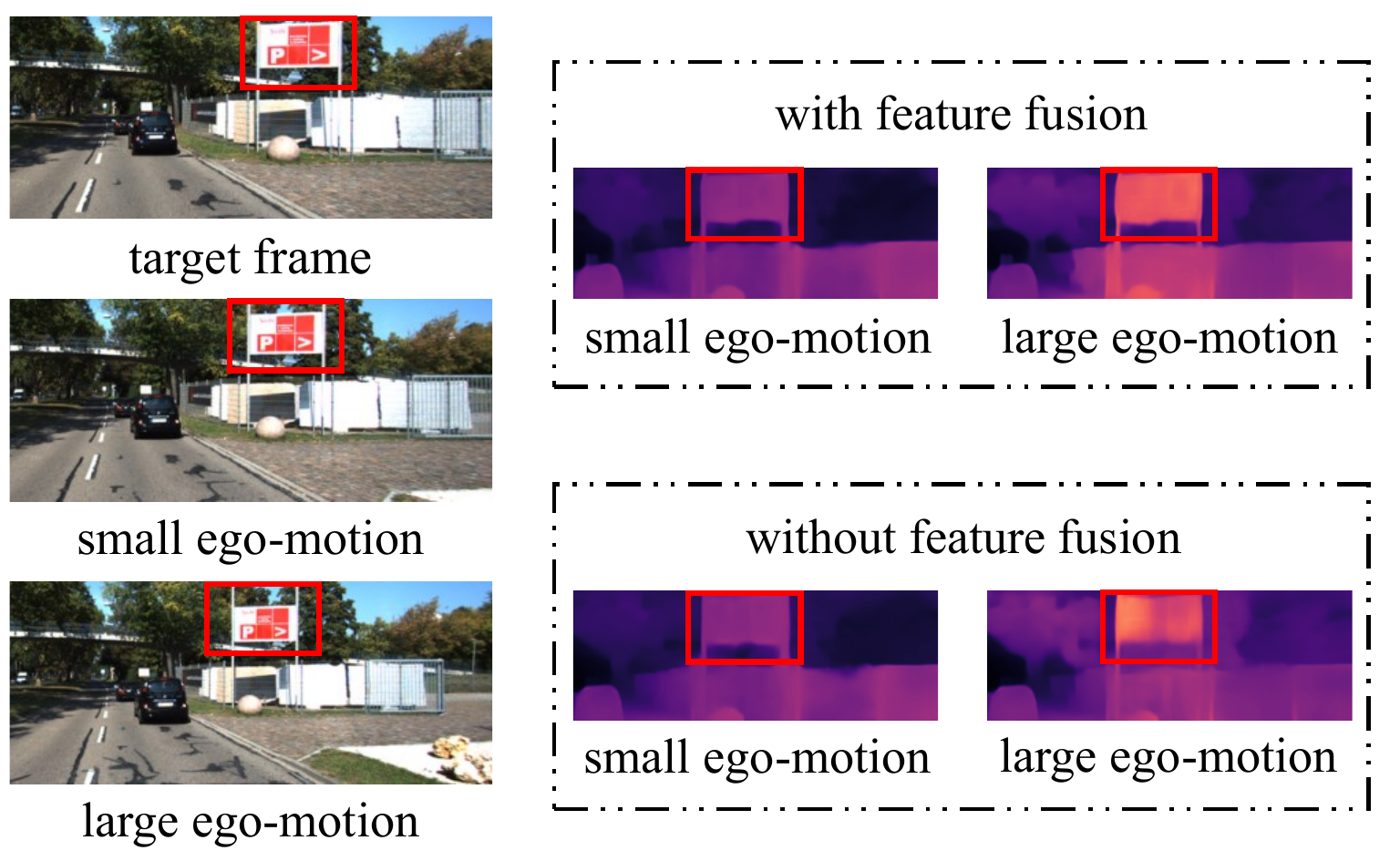}
    \caption{Depth estimation with/without multi-scale feature fusion in different camera ego-motion. }
    \label{fig:largemotion}
\end{figure}

As shown in Figure~\ref{fig:largemotion}, due to the ego-motion of the camera view, the feature points to be matched are in different scales in adjacent frames. For instance, when the camera is approaching the objects, the scale of the objects in the image gradually becomes larger. To enhance the characterization ability of texture features and the accuracy of feature matching, the multi-scale feature fusion block is proposed. Different from previous works~\cite{Guo_2020_CVPR}, we pursue finer-grained  multi-scale feature representations that preserve differences in pixel-level features, rather than hoping that the multi-scale features are embedded in the entire feature map. 
Given a small-scale feature $F_1$ and a middle-scale feature $F_2$ from feature encoder $N_e$, the fused features are generated by:
\begin{equation}
    \begin{array}{l}
    F_{12} = Convs(Convs(F_1) \oplus F_2),\\
    F_{32} = Up.(ResBlock(F_2)), \\
    F_{ms} = F_2 \oplus F_{12} \oplus F_{32},\\
    \end{array}    
\end{equation}
where $Up.(\cdot)$ is the bilinear upsample operator, $\oplus $ is the concatenation operator, and $ResBlock(\cdot)$ are the residual blocks of the third layer in ResNet-18~\cite{he2016deep}. 
Inspired by the multi-head mechanism~\cite{guizilini2022multi}, we aggregate features of different scales in different groups explicitly. Small-scale features can focus on the texture details of the image. Large-scale features that have larger receptive fields are more friendly to low-textured regions. 
As shown in Figure~\ref{fig:largemotion}, our multi-scale feature fusion improves the depth estimation with large camera ego-motion. Quantitative results in Table~\ref{table:speed} also show the effectiveness of the fusion block.

\subsection{Learn from inconsistent depth}
 \label{sec:overfitting}

\subsubsection{How the depth becomes inconsistent in dynamic regions}

\begin{figure}[t]
    \centering
    \includegraphics[width=0.8\linewidth]{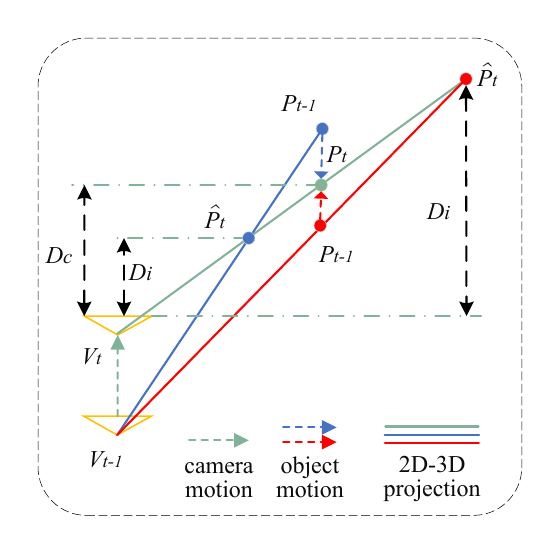}
    \caption{The depth inconsistency problem of the dynamic objects. $V_t$ and $V_{t-1}$: camera views at time $t$ and $t-1$. $P_t$ and $P_{t-1}$: the 3D coordinates of the dynamic objects at time $t$ and $t-1$. $D_c$ and $D_i$: consistent depth prediction and inconsistent depth prediction. $\hat{P}_t$: predicted 3D coordinates from $D_i$.
    }
    \label{fig:overfit}
\end{figure}

In the photo-metric loss or the cost volume in the multi-frame depth network, the frame $I_s$ or the feature $F_s$ is warped to the target view through the re-projection function under the assumption of a static environment. 
As the green line in Figure~\ref{fig:overfit} shows, given the depth prediction $D_c$, camera intrinsic $K$, and the pixel coordinate $[u_t,v_t]$, the 3D coordinate $P_t$ of the corresponding point can be obtained through the pinhole camera geometry~\cite{hartley2003multiple}:
\begin{equation}
    P_t = D_cK^{-1}[u_t,v_t,1]^T,
    \label{eqn:projs}
\end{equation}
where $P_t$ is the 3D coordinate from the view $V_t$. With the pose estimation $T_{t \rightarrow t-1}$ and the motion $T_{obj}$ of the corresponding objects, the 3D coordinate $\hat{P}_{ t-1}$ from the view $V_{t-1}$ is calculated by:
\begin{equation}
    \hat{P}_{t-1} = T_{t \rightarrow t-1} P_t + T_{obj}.
    \label{eqn:dymotion}
\end{equation}
 Then the corresponding pixel $[\hat{u}_{t-1},\hat{v}_{t-1}]$ at time $t-1$ of $[u_t,v_t]$ can be obtained by:
\begin{equation}
    [\hat{u}_{t-1},\hat{v}_{t-1}, 1] ^ T = K \hat{P}_{ t-1}.
    \label{eqn:reproj}
\end{equation}
Under the static scene assumption, the object motions $T_{obj}$ are all considered as zeros, which is true in most regions. In static regions, the consistent depth prediction $D_c$ will lead to the right corresponding pixel in adjacent frames resulting in low self-supervise losses.
However, in the dynamic regions, minimizing the plain self-supervise loss will lead to inconsistent depth prediction $D_i$. As the blue/red line in Figure~\ref{fig:overfit} shows, once the projection point $\hat{P}_t$ from inconsistent depth $D_i$ is on the line $\overline{V_{t-1}P_{t-1}}$, the same matching pixel coordinate $[\hat{u}_{t-1},\hat{v}_{t-1}]$ can be obtained since the camera projection model is a linear function. 
As illustrated in Figure~\ref{fig:overfit}, the inconsistent depth will be greater or smaller than the ground truth according to the motion pattern of the dynamic objects. 

\begin{figure*}
    \centering
    \includegraphics[width=1.0\linewidth]{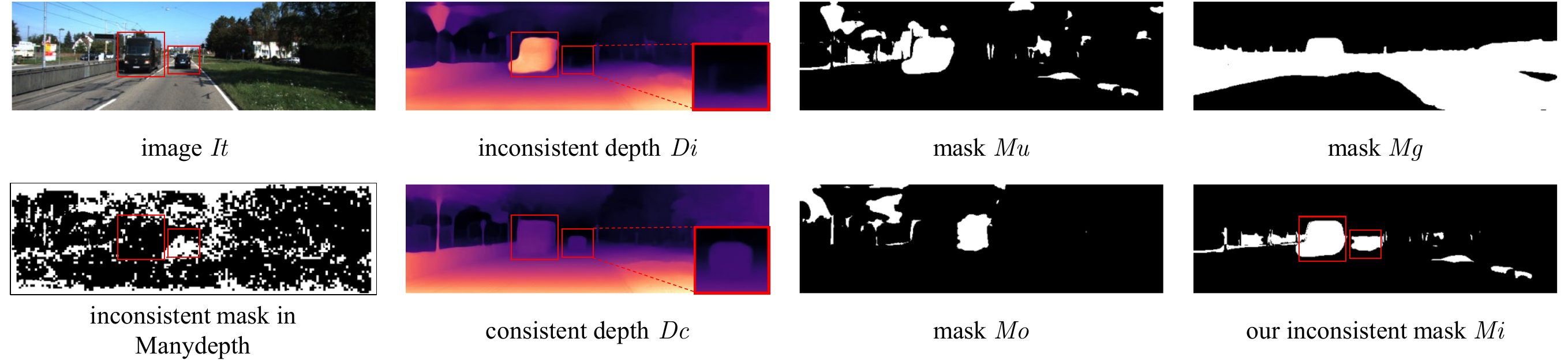}
    \caption{Intermediate results of our depth inconsistency mask $M_i$ and the comparison with previous inconsistency mask proposed by ManyDepth\cite{watson2021temporal}. Masks $M_u$ show the under-estimation regions of the inconsistent depth, which usually implies the objects with the opposite direction of movement to the camera. Masks $M_o$ show the over-estimation regions of the inconsistent depth, which usually implies the objects with the same direction of movement as the camera. Ground masks $M_g$ are used to filter out some rigid regions.
    }
    \label{fig:masks}
\end{figure*}

\subsubsection{Generating depth inconsistency mask}

As mentioned above, training the depth network by minimizing the photo-metric loss function under the static environment assumption will lead to under/over-estimation problems in dynamic regions. To relieve the problem of dynamic objects, we leverage the inconsistent performance of the multi-frame depth network to generate the depth inconsistency mask. Due to the embedded cost volume, the under/over-estimation problem of the multi-frame depth network is much heavier than that of the single-frame depth network, since the single-frame texture feature are similar whether the objects are dynamic or static but the spatial-temporal feature will be much different in these two different patterns.

Under this observation, we first train a multi-frame depth network without any regularization on the dynamic regions.
In order to keep the consistent performance in static regions, we use the static frame augmentation and zero cost volume augmentation~\cite{watson2021temporal} during training, such that the 
multi-frame depth network can learn from the texture features rather than rely on the spatial-temporal features. The pre-trained multi-frame network is then used to generate inconsistent depth prediction $D_i$. To obtain consistent depth prediction $D_i$, we pre-train a single-frame depth network separately as~\cite{godard2019digging}. 
For more consistent performance in dynamic regions, we use the HRNet-18~\cite{wang2020deep} as the backbone of the single-frame depth network. The dense feature fusion in HRNet enhances the robustness of texture feature extraction, leading to robust performance in dynamic regions. Given the inconsistent depth prediction $D_i$ and consistent depth prediction $D_c$, we express the depth inconsistency masks from $M_o$, $M_u$, and $M_g$:
\begin{equation}
    \begin{array}{l}
    M_o = [  D_i^{\ast} > \alpha D_c],\\
    M_u = [  D_i^{\ast} < \beta D_c], \\
    M_g = [- y_g< \hat{Y} < y_g],\\
    \end{array}    
\end{equation}
where $M_o$ is the over-estimation mask, $M_u$ is the under-estimation mask, $D_i^{\ast}$ is the scale aligned depth by median value aligning between $D_i$ and $D_c$, $y_g$ is the camera height calculated by fitting the ground plane~\cite{xue2020toward} and $\hat{Y}$ is the $y$-coordinate of the corresponding pixel. The parameters $\alpha$ and $\beta$ are two thresholds set to detect the regions of the over/under-estimation.
As shown in Figure~\ref{fig:masks}, due to the difference in network architecture and training procedure, the inconsistent depth $D_i$ and consistent depth $D_c$ may disagree with each other even in some static regions, leading to wrong dynamic detection. Noting that dynamic objects like cars and pedestrians are on the ground, we use the ground mask $M_g$ to filter out the false positive dynamic regions. The final depth inconsistency binary mask $M_i$ is calculated by:
\begin{equation}
    M_i = (M_o \odot M_u) \bullet M_g,
\end{equation}
where $\odot$ is the logical or operator and $\bullet$ is the logical and operator. The depth inconsistency mask is set to $1$ in dynamic regions and $0$ in static regions. 
As shown in Figure~\ref{fig:masks}, compared to the inconsistency mask proposed by ManyDepth~\cite{watson2021temporal}, our depth inconsistency masks $M_i$ can better filter out dynamic regions and keep more static regions. The depth inconsistency masks $M_i$ are used to filter out the dynamic regions during training.

\subsection{Robust knowledge distillation}
\label{sec:robust}

Although the inconsistency masks can filter out the dynamic regions in the self-supervise loss, the multi-frame depth network may still obtain inconsistent depth estimation in dynamic regions~\cite{watson2021temporal}. Since the embedded spatial-temporal features are quite different in the dynamic regions, which means that the multi-frame depth network requires additional regularization losses for the dynamic regions. Following~\cite{watson2021temporal}, we use a single-frame depth network as the teacher network for knowledge distillation.
As discussed above, using the HRNet-18~\cite{wang2020deep} as the backbone of the single-frame depth network provides more consistent performance in dynamic regions. Therefore, we also choose the HRNet-18~\cite{wang2020deep} as the backbone of the teacher network for more robust knowledge distillation. The knowledge distillation loss is calculated by:
\begin{equation}
  L_c = \sum M_i | \hat{D}_t - \hat{D}_{t,s} |,
  \label{eqn:kdloss}
\end{equation}
where $\hat{D}_{t}$ is the depth estimation of multi-frame depth network, $\hat{D}_{t,s}$ is the depth estimation of the teacher network, and $M_i$ is our consistency mask. The teacher network is trained jointly with the multi-frame depth network and the pose network. Since the consistent depth estimation in the dynamic regions will lead to large self-supervise loss, which may confuse the learning of the pose network, the inconsistency masks $M_i$ are also used for the training of the teacher network.

\begin{table*}
    \caption{The comparison of depth estimation with other monocular self-supervised methods on KITTI and Cityscapes datasets.  \emph{Test frames}: input frames of the depth network during testing. \emph{Semantic}: whether additional semantic information was used during training. $W\times H$: the resolution of input frames during testing. Following ManyDepth, we use the “A" crop for Cityscapes. The best results are in bold, second best results are underlined.
     }
    \vspace{-8pt}
	\label{table:results}
	\begin{center}
		\resizebox{1.0\textwidth}{!}
		{
			\begin{tabular}{|c|l|c|c|c|cccc|ccc|}
                \hline
				& \multirow{2}{*}{\bf Method} & \multirow{2}{*}{Test frames} & \multirow{2}{*}{Semantic} & \multirow{2}{*}{$W\times H$}   & AbsRel & SqRel & RMSE & RMSE log & $\delta < 1.25$ & $\delta < 1.25^2$ &  $\delta < 1.25^3$\\
				& & & & & \multicolumn{4}{c|}{$Lower\ is\ better$} & \multicolumn{3}{c|}{$Higher\ is\ better$}\\
				\hline
                \hline
                
                & Monodepth2~\cite{godard2019digging} & 1 & & $640 \times 192$ & 0.115  &   0.903   &  4.863  &   0.193   &  0.877  &   0.959  &   0.981\\
                & Lee et al.~\cite{lee2021attentive} & 1  & $\surd$ & $832 \times 256$ & 0.114   &   0.876   &   4.715   &   0.191   &   0.872   &   0.955   &   0.981\\
                & DiPE~\cite{jiang2020dipe} & 1 & & $640 \times 192$   &   0.112   &   0.875   &   4.795   &   0.190   &   0.880   &   0.960   &   0.981\\
                & InstaDM~\cite{lee2021learning}  & 1  & $\surd$ & $832 \times 256$ & 0.112   &   0.777   &   4.772   &   0.191   &   0.872   &   0.959   &   0.982\\
                &Patil et al.~\cite{patil2020don} & N & & $640 \times 192$ &  0.111   &   0.821   &   4.650   &   0.187   &   0.883   &   0.961   &   0.982\\
                &Packnet-SFM~\cite{guizilini20203d} & 1 & & $640 \times 192$ & 0.111   &   0.785   &   4.601   &   0.189   &   0.878   &   0.960   &   0.982\\
                &Wang et al.~\cite{wang20223d}& 1  &   & $832 \times 256$ &0.109   &   0.790   &   4.656   &   0.185   &   0.882   &   0.962   &   0.983\\
                &RM-depth~\cite{hui2022rm}& 1 & & $640 \times 192$ & 0.108   &   0.710   &   4.513   &   0.183   &   0.884   &   0.964   &   0.983\\
                &Johnston et al.~\cite{johnston2020self} & 1 & & $640 \times 192$ & 0.106   &   0.861   &   4.699   &   0.185   &   0.889   &   0.962   &   0.982\\
                &Wang et al.~\cite{wang2020self}& 2 (-1, 0) &          & $640 \times 192$ &0.106   &   0.799   &   4.662   &   0.187   &   0.889   &   0.961   &   0.982\\
                &TC-depth~\cite{ruhkamp2021attention}& 3 (-1, 0, +1) &          & $640 \times 192$ &0.103   &    0.746   &    4.483   &    0.180   &    0.894   &    0.965   &    0.983\\
                &DIFFNet~\cite{zhou2021self} & 1  &    & $640 \times 192$ & 0.102   &    0.764   &    4.483   &    0.180   &    0.896   &    0.965   &    0.983\\
                &Guizilini et al.~\cite{guizilini2020semantically} & 1  & $\surd$ & $640 \times 192$ & 0.102   &   0.698   &   4.381   &   0.178   &   0.896   &   0.964   &   \underline{0.984}\\
				& ManyDepth~\cite{watson2021temporal}    & 2 (-1, 0) &          & $640 \times 192$ & 0.098     &   0.770    &    4.459    &    0.176    &    0.900    &    0.965    &    0.983\\
				& Dynamicdepth~\cite{feng2022disentangling} & 2 (-1, 0) & $\surd$  & $640 \times 192$ & \underline{0.096}    &    0.720    &    4.458    &    0.175    &    0.897    &    0.964    &    \underline{0.984}\\
                &RA-depth~\cite{he2022ra} & 1  &   & $640 \times 192$ & \underline{0.096}    &    \bf0.632    &    4.216    &    \underline{0.171}    &    0.903    &    \bf0.968    &    \bf0.985\\
				& DepthFormer~\cite{guizilini2022multi}  & 2 (-1, 0) &          & $640 \times 192$ & \bf0.090    &     0.661    &     \bf4.149    &     0.175    &     \underline{0.905}    &     \underline{0.967}    &     \underline{0.984}\\
                \cline{2-12}
                  \multirow{-17}{*}{\rotatebox{90}{KITTI}} &  \bf Ours     & 2 (-1, 0) &          & $640 \times 192$ &   \bf0.090  &   \underline{0.653}  &   \underline{4.173}  &   \bf0.167  &   \bf0.911  &   \bf0.968  &   \bf0.985  \\
                \hline
                \hline
                &Li et al.~\cite{li2021unsupervised} & 1  &   & $416 \times 128$ & 0.119   &    1.290   &    6.980   &    0.190   &    0.846   &    0.952   &    0.982\\
                & Lee et al.~\cite{lee2021attentive} & 1  & $\surd$ & $832 \times 256$ &0.116   &    1.213   &    6.695   &    0.186   &    0.852   &    0.951   &    0.982\\
                & InstaDM~\cite{lee2021learning}  & 1  & $\surd$ & $832 \times 256$ &0.111   &    1.158   &    6.437   &    0.182   &    0.868   &    0.961   &    0.983\\
                & ManyDepth~\cite{watson2021temporal}    & 2 (-1, 0) &          &    $416 \times 128$ &    0.114   &   1.193   &   6.223   &   0.170   &   0.875   &   0.967   &   0.989\\
                & DynamicDepth~\cite{feng2022disentangling} & 2 (-1, 0) &  $\surd$ &   $416 \times 128$  &   0.103   &    1.000    &   5.867    &   0.157   &    \underline{0.895}   &    0.974    &   0.991\\
                & RM-depth~\cite{hui2022rm}     &     1     &          &  $640 \times 192$   &   \underline{0.100}    &  \bf0.839    &   \underline{5.774}    &   \underline{0.154}    &   \underline{0.895}    &   \underline{0.976}    &   \bf0.993 \\
                \cline{2-12}
                 \multirow{-7}{*}{\rotatebox{90}{Cityscapes}} & \bf Ours      & 2 (-1, 0) &          &  $416 \times 128$   &   \bf0.098  &  \underline {0.946}  &   \bf5.553  &   \bf0.148  &   \bf0.908  &   \bf0.977  &   \underline{0.992}  \\
                \hline
			\end{tabular}
        }
	\end{center}

\end{table*}

In order to strengthen the texture feature which provides consistent features for dynamic objects and static objects, we adopt the static frame augmentation and zero cost volume augmentation as~\cite{watson2021temporal}. With these two types of augmentation, the cost volume fails to provide helpful spatial-temporal features, and the multi-frame depth network will be forced to predict depth only from the texture features. Different from considering all the pixels as the inconsistent pixels for regularization as~\cite{watson2021temporal},
we check the reliability of the depth pseudo supervision between the teacher depth and the multi-frame depth for robust knowledge distillation. Given the depth predictions, the robust mask $M_r$ is calculated by:
\begin{equation}
    M_r = [ L_{ph,s} < L_{ph}],
\end{equation}
where $L_{ph,s}$ and $L_{ph}$ are the photo-metric loss of the single-frame depth and multi-frame depth, respectively, and $[\cdot ]$ is the Iverson bracket. The robust mask filters out the unreliable depth supervision from the teacher network and thus provides robust knowledge distillation for the multi-frame depth network. For the augmented samples, the consistency loss is:
\begin{equation}
    L_c = \sum (M_r \bullet (1 - M_i))|  \hat{D}_t - \hat{D}_{t,s}  |,
\end{equation}
where $\bullet$ is the logical and operator, and $M_i$ is the inconsistency mask. The consistency mask $1-M_i$ is used to filter out the dynamic regions since the lower photo-metric loss means inconsistent depth in dynamic regions. When without augmentation, we use the plain knowledge loss as Eq.~\ref{eqn:kdloss} for the regularization in dynamic regions.

The final loss is:
\begin{equation}
    L= (1-M_i)L_{ph} + L_c + \beta L_{sm} + (1-M_i)L_{ph,s} + \beta L_{sm,s},
    \label{eqn:totalloss}
\end{equation}
where $L_{ph,s}$ and $L_{sm,s}$ are losses of the teacher network and $\beta = 1e^{-3}$. The teacher network is trained jointly throughout the training procedure.

\section{Experiments}

\subsection{Datasets}

\subsubsection{KITTI}
The KITTI~\cite{geiger2012we} is a standard depth evaluation benchmark. To compare with other approaches, we use the data split of Eigen et al.~\cite{eigen2015predicting}. Following ManyDepth~\cite{watson2021temporal}, we adopt Zhou et al.'s~\cite{zhou2017unsupervised} pre-processing to remove the static frames. This results in 39810/4424/697 monocular samples for training/validation/testing.

\subsubsection{Cityscapes}
Cityscapes~\cite{cordts2016Cityscapes} is a challenging dataset full of dynamic scenes. Following ManyDepth~\cite{watson2021temporal}, we use 69731 monocular samples for training. For testing, we use the official testing set that contains 1525 samples with the SGM~\cite{hirschmuller2007stereo} disparity maps.

\subsection{Implementation details}

\subsubsection{Training details}
We implement our work in Pytorch~\cite{paszke2019pytorch}. During training, we use color and flip augmentations on the images~\cite{godard2019digging}. We train the networks with Adam~\cite{kingma2014adam} for 20 epochs with the learning rate $1e^{-4}$ for the first 15 epochs and $1e^{-5}$ for the last 5 epochs. Different from ManyDepth~\cite{watson2021temporal}, we do not freeze the teacher network and pose network for all 20 epochs for KITTI~\cite{geiger2012we}. The batch size is 6 for KITTI. But for Cityscapes~\cite{cordts2016Cityscapes}, we freeze the teacher network and the pose network after 112000 training samples (14000 steps with batch size as 8) and de-freeze the teacher network and the pose network for the last 5 epochs. We find it important in this challenging dataset which contains more dynamic scenes and training samples than the KITTI dataset. The hypothesized depth range is fixed when the teacher network and pose network are frozen. Depth bins $D$ are set as $96$ and the hypothesized depth range is iteratively adapted from the teacher depth as ~\cite{watson2021temporal}. All experiments are performed on a single Nvidia RTX 3080Ti GPU within 11 GB memory.

\subsubsection{Networks}
As mentioned above, we use HRNet-18~\cite{wang2020deep} as the encoder of the consistent teacher. For the ablation study, we use ResNet-18~\cite{he2016deep} as the encoder of the teacher as ManyDepth~\cite{watson2021temporal} in the case without the consistent teacher. We use ResNet-18~\cite{he2016deep} as the backbone of the multi-frame depth encoder. We use the depth decoder as Monodepth2~\cite{godard2019digging} and output four scales depth for the depth network with ResNet-18~\cite{he2016deep} as the encoder. Nevertheless, we only output single-scale depth for training efficiency, when using HRNet-18~\cite{wang2020deep} as the depth encoder. The pose network is the same as Monodepth2~\cite{godard2019digging}.

\begin{figure*}[t]
    \centering
    \includegraphics[width=0.9\linewidth]{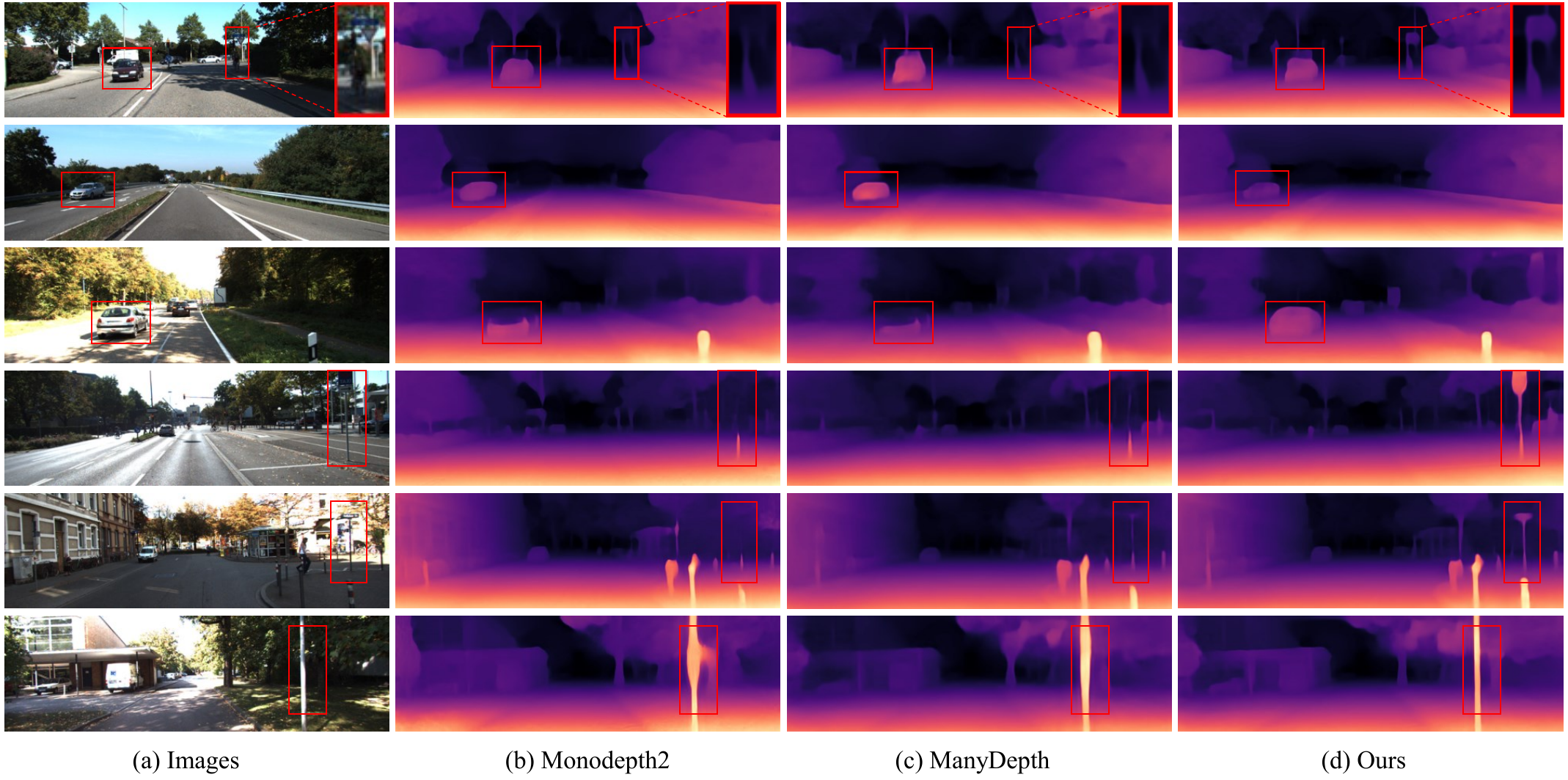}
    \vspace{-8pt}
    \caption{Qualitative depth estimation comparison with previous methods. 
    }
    \vspace{-8pt}
    \label{fig:visual}
\end{figure*}

\subsection{Depth evaluation results}

Following the previous monocular self-supervised works~\cite{watson2021temporal,feng2022disentangling}, we use the standard evaluation errors as metrics, perform scale alignment between the depth prediction and ground truth, and evaluate only less than 80 meters. The listed results of KITTI are all evaluated on origin depth from velodyne. The results of Cityscapes are evaluated on the official test set. Table~\ref{table:results} and Figure~\ref{fig:visual} show the comparison between the previous methods and our method. We rank methods based on the absolute-relative-error. Our method achieves state-of-the-art (SOTA) results compared to previous methods on both KITTI and Cityscapes datasets. Our method outperforms the previous works on both single-frame and multi-frame methods. Compared to the previous semantic-based methods~\cite{lee2021attentive,lee2021learning,feng2022disentangling}, we do not need semantic labels or pre-trained semantic segmentation model which leverages extra semantic labels. Different from the motion model-based methods~\cite{li2021unsupervised,hui2022rm}, our depth inconsistency masks are off-the-shelf once generated, which saves the training time cost on the motion prediction networks. Our multi-scale feature fusion provides more accurate depth cues and is more efficient than the transformer-based method~\cite{guizilini2022multi}. Our method needs only 3.8 GB GPU test memory, while DepthFormer~\cite{guizilini2022multi} needs 6.4 GB GPU test memory. In the challenging Cityscapes dataset, our method outperforms the previous multi-frame method with auxiliary semantic information and single depth prior~\cite{feng2022disentangling}. Our method achieves better results than the SOTA single-frame method~\cite{hui2022rm}.
As shown in Figure~\ref{fig:visual}, our depth model predicts more consistent depth in dynamic regions and performs better on small-scale objects and the edges.

\setlength{\tabcolsep}{0.9mm}
\begin{table*}[t]
    \renewcommand\arraystretch{1.125}
 	\caption{The ablation results of KITTI and Cityscapes. K$\ast$ is the result of the consistent teacher network on KITTI.}
	\begin{center}
		\resizebox{0.85\textwidth}{!}
		{
			\begin{tabular}{|c|c|c|c|c|cccc|ccc|}
                \hline
				& Depth inconsistency & Multi-scale& \multicolumn{2}{c|}{Robust knowledge distillation} & AbsRel & SqRel & RMSE & RMSE log & $\delta < 1.25$ & $\delta < 1.25^2$ &  $\delta < 1.25^3$\\
				\cline{4-12}
				& mask  &  fusion & Consistent teacher & Robust mask & \multicolumn{4}{c|}{$Lower\ is\ better$} & \multicolumn{3}{c|}{$Higher\ is\ better$}\\
				\hline
                \hline
                &  & & &  &   0.106  &   0.788  &   4.527  &   0.183  &   0.893  &   0.964  &   0.983  \\
                \multirow{-2}{*}{\rotatebox{90}{K$\ast$}} &  $\surd$ & & &   &  \bf0.101  &   \bf0.738  &   \bf4.435  &   \bf0.178  &   \bf0.896  &   \bf0.965  &   \bf0.984  \\
                \hline
                \hline
                
				&  & & & & 0.098     &   0.770    &    4.459    &    0.176    &    0.900    &    0.965    &    0.983\\
                &  & $\surd$ & &   &  0.096  &   0.752  &   4.384  &   0.176  &   0.903  &   0.965  &   0.983  \\
				&  $\surd$ & & &   & 0.096  &   0.735  &   4.369  &   0.173  &   0.902  &   0.966  &   0.984  \\
                & & & $\surd$ &   & 0.096  &   0.697  &   4.275  &   0.174  &   0.903  &   0.966  &   0.984  \\
                &  & & $\surd$ & $\surd$  &   0.094  &   0.719  &   4.288  &   0.171  &   0.905  &   0.967  &   0.984  \\
                \multirow{-6}{*}{\rotatebox{90}{KITTI}}&  $\surd$ & $\surd$ & $\surd$  &$\surd$ & \bf0.090  &   \bf0.653  &   \bf4.173  &   \bf0.167  &   \bf0.911  &   \bf0.968  &   \bf0.985  \\ 
                \hline
                \hline
                
                &   & & &  &  0.114  &   1.274  &   6.188  &   0.170  &   0.882  &   0.967  &   0.988  \\
                & $\surd$  & &  &  &   0.103  &   1.092  &   5.845  &   0.157  &   0.897  &   0.973  &   0.990  \\
				&  $\surd$ &  & $\surd$ & $\surd$  & \bf0.098  &   0.970  &   5.631  &   0.149  &   0.905  &   \bf0.977  &   \bf0.992  \\
                \multirow{-4}{*}{\rotatebox{90}{Cityscapes}}&  $\surd$ & $\surd$ & $\surd$ & $\surd$  & \bf0.098  &  \bf0.946  &   \bf5.553  &   \bf0.148  &   \bf0.908  &   \bf0.977  &   \bf0.992  \\
                \hline
			\end{tabular}
	   }
	\end{center}
	\label{table:ablation}
\end{table*}

\subsection{Ablation study}
\label{sec:ablation}
We perform ablation experiments on both the KITTI dataset and the Cityscapes dataset. The results are as shown in Table~\ref{table:ablation}. We use the depth evaluation results trained in our proposed freezing way for the baseline of Cityscapes, rather than the direct results from ManyDepth~\cite{watson2021temporal} to show the effectiveness of our method clearly.

\subsubsection{Multi-scale feature fusion} 
As shown in Table~\ref{table:ablation}, the multi-scale feature fusion block improves the depth evaluation results. Compared to the experiments on Cityscapes~\cite{cordts2016Cityscapes}, the multi-scale feature fusion block achieves larger improvement on KITTI~\cite{geiger2012we}. This is due to the fact that the camera ego-motions are smaller between adjacent frames in the Cityscapes dataset (camera frequency 17 Hz) than in the KITTI dataset (camera frequency 10 Hz), which implies a smaller scale difference in the feature points between adjacent frames in the Cityscapes dataset.
We evaluate the depth on the Cityscapes dataset with different test frames as inputs, as shown in Table~\ref{table:speed}. The results demonstrate the effectiveness and robustness of the multi-scale feature fusion block in the presence of large camera ego-motion between adjacent frames.

\setlength{\tabcolsep}{0.9mm}
\begin{table}[h]
	\centering
        \caption{The depth evaluation results on the Cityscapes dataset with different test frames.  The longer time interval between adjacent frames implies larger camera ego-motions.}
	\vspace{-2pt}	
    \renewcommand\arraystretch{1.125}
	\footnotesize
	\begin{center}
		\resizebox{0.475\textwidth}{!}
		{
			\begin{tabular}{|c|c|ccc|cc|}
                \hline
                Test & Multi-scale & AbsRel & RMSE & RMSE log & $\delta < 1.25$ & $\delta < 1.25^2$ \\
				\cline{3-7}
                Frames &  Fusion & \multicolumn{3}{c|}{$Lower\ is\ better$} & \multicolumn{2}{c|}{$Higher\ is\ better$}\\
                \hline
               \hline
                & 
                &   \bf0.098  &   5.631 &   0.149  &   0.905  &   \bf0.977   \\
                \multirow{-2}{*}{$I_t,I_{t-1}$}&  $\surd$     
                &   \bf0.098   &   \bf5.553  &   \bf0.148  &   \bf0.908  &   \bf0.977   \\
				\hline
                \hline
                & 
                &   0.101   &   5.670  &   0.152  &   0.903  &   0.976   \\
                \multirow{-2}{*}{$I_t,I_{t-2}$}&  $\surd$  
                &   \bf0.099   &   \bf5.566  &   \bf0.149  &   \bf0.906  &   \bf0.977  \\
				\hline
                \hline
                & 
                &   0.104   &   5.755  &   0.155  &   0.898  &   0.975   \\
                \multirow{-2}{*}{$I_t,I_{t-3}$}& $\surd$  
                &   \bf0.101   &   \bf5.625  &   \bf0.152  &   \bf0.903  &   \bf0.976   \\
                \hline
			\end{tabular}
	   }
	\end{center}
	\vspace{-2pt}	
	\label{table:speed}
\end{table}

\setlength{\tabcolsep}{0.9mm}
\begin{table}[h]
	\centering
        \caption{The depth evaluation results of the dynamic objects (e.g. vehicles, bikes, and pedestrians) on Cityscapes dataset. Dynamic object masks are generated by pre-trained model EffcientPS~\cite{mohan2021efficientps}. $M_i$: depth inconsistency mask, $M_r$: robust knowledge distillation.}
	\vspace{-2pt}	
    \renewcommand\arraystretch{1.125}
	\footnotesize
	\begin{center}
		\resizebox{0.475\textwidth}{!}
		{
			\begin{tabular}{|l|ccc|cc|}
                \hline
                \multirow{2}{*}{Methods}  & AbsRel  & RMSE & RMSE log & $\delta < 1.25$ & $\delta < 1.25^2$ \\
				\cline{2-6}
                  & \multicolumn{3}{c|}{$Lower\ is\ better$} & \multicolumn{2}{c|}{$Higher\ is\ better$}\\
                \hline
               \hline
                ManyDepth~\cite{watson2021temporal} &   0.172    &   5.829  &   0.214  &   0.792  &   0.917  \\
                DynamicDepth~\cite{feng2022disentangling} &   0.145    &   5.143  &   0.183  &   0.830  &   0.950  \\
				\hline
                \hline
                Baseline &   0.178  &   6.123  &   0.222  &   0.787  &   0.911  \\
                Baseline + $M_i$ &   0.143   &   5.151  &   0.185  &   0.847  &   0.951   \\
                Baseline + $M_i$ + $M_r$ &  \bf 0.127   &  \bf 4.717  &  \bf 0.169  &  \bf 0.862  &  \bf 0.961 \\
                \hline
			\end{tabular}
	   }
	\end{center}
	\vspace{-2pt}	
	\label{table:dynamic}
\end{table}

\subsubsection{Depth inconsistency mask}

As shown in Tables~\ref{table:ablation} and ~\ref{table:dynamic},
the results illustrate the effectiveness of our proposed depth inconsistency masks. Using our depth inconsistency masks to filter out the dynamic objects can improve the depth evaluation results, regardless of whether a multi-scale feature fusion block or robust knowledge distillation is adopted. The depth inconsistency masks improve the performance of the teacher network as shown in Table~\ref{table:ablation}. As mentioned in Sec.~\ref{sec:robust}, the robust depth prediction in dynamic regions will mislead the pose network through the photo-metric loss. Therefore, using the depth inconsistency masks to filter out the dynamic regions during training improves the pose prediction, which in turn improves the single-frame depth prediction. We use the threshold parameters as $\alpha=2$ and $\beta=0.85$ for both KITTI and Cityscapes.

\subsubsection{Robust knowledge distillation} 
As shown in Tables~\ref{table:ablation} and ~\ref{table:dynamic}, the robust knowledge distillation improves the multi-frame depth evaluation results. Although using the robust knowledge distillation extends the training time from about eight hours to about eleven hours, the inference speed as well as the memory usage of the multi-frame depth network during testing are not affected, since the structure of the multi-frame depth network is not changed.

\section{Conclusion}

In this paper, we propose a novel self-supervised multi-frame depth learning framework. The multi-frame depth network utilizes multi-scale feature fusion for texture feature extraction and feature matching. Different from previous methods that directly regularize the inconsistent regions during training, we propose to first train an inconsistent depth model and then generate the depth inconsistency masks off-the-shelf. The depth inconsistency mask which provides information about dynamic objects, improves the performance of the depth network in dynamic scenes. The robust knowledge distillation improves the multi-frame depth with no increase in inference cost. With these contributions, our method achieves SOTA performance on the KITTI and Cityscapes datasets.

The limitation is that our method needs to pre-train the consistent single frame network and inconsistent multi-frame network to calculate the depth inconsistency masks, which will increase the training time. Saving these binary-masks will demand extra storage space depending on the format. Calculating the masks online will decrease the training efficiency. We will discuss this further in future research.

\begin{acks}
This work was supported by the National Key R\&D Program of China under Grant 2021ZD0112600, the National Natural Science Foundation of China under Grants 62173283 and 61977046, Shanghai Municipal Science and Technology Major Project (No. 2021SHZD ZX0102) and the Shanghai Science and Technology
Program under Grant (No. 21JC1400600).
\end{acks}

\bibliographystyle{ACM-Reference-Format}
\bibliography{1542_full_paper_ref}

\appendix
\newpage
\section{Evaluation Metrics}
Following Eigen et al.~\cite{eigen2015predicting}, we use several error metrics and accuracy metrics to show the performance of the depth:
\begin{itemize}
    \item Relative Absolute Error (\textbf{AbsRel}): 
    
    $\frac{1}{N}\sum_{i,j}|\hat{D}^{i,j}-D^{i,j}_{gt}|/D^{i,j}_{gt}$;
    \item Relative Squared Error (\textbf{SqRel}):

    $\frac{1}{N}\sum_{i,j}(\hat{D}^{i,j}-D^{i,j}_{gt})^2/D^{i,j}_{gt}$;
    \item Root Mean Squared Error (\textbf{RMSE}):

    $\frac{1}{N}\sum_{i,j}\sqrt{(\hat{D}^{i,j}-D^{i,j}_{gt})^2}$;
    \item Root Mean Squared Logarithmic Error (\textbf{RMSE log}):

    $\frac{1}{N}\sum_{i,j}\sqrt{(log\hat{D}^{i,j}-logD^{i,j}_{gt})^2}$;

    \item threshold accuracy (\boldmath $\delta < 1.25^k$ \unboldmath): percentage of $\hat{D}^{i,j}$ 
    
    s.t. $max(\frac{\hat{D}^{i,j}}{D^{i,j}_{gt}},\frac{D^{i,j}_{gt}}{\hat{D}^{i,j}}) < 1.25^k$.
\end{itemize}

\section{Quantitative Comparison}

\subsection{Depth evaluation on KITTI improved ground truth}
As shown in Table~\ref{table:resultsapp}, we show more results for comparison. The depth results are evaluated on the improved depth maps from~\cite{uhrig2017sparsity} on the KITTI Eigen test split~\cite{eigen2015predicting}, for distances up to 80m with the Garg crop~\cite{garg2016unsupervised}. Our method also achieves good performance compared to previous works on KITTI improved ground truth.

\begin{table}[h]
    \caption{The comparison of depth estimation on KITTI improved ground truth.}	
	\label{table:resultsapp}
    \vspace{-6pt}
		\resizebox{0.475\textwidth}{!}
		{
			\begin{tabular}{|l|cccc|ccc|}
                \hline
				\multirow{2}{*}{\bf Method}  & AbsRel & SqRel & RMSE & RMSE log & $\delta < 1.25$ & $\delta < 1.25^2$ &  $\delta < 1.25^3$\\
                \cline{2-8}
				 & \multicolumn{4}{c|}{$Lower\ is\ better$} & \multicolumn{3}{c|}{$Higher\ is\ better$}\\
				\hline
                \hline
                Monodepth2~\cite{godard2019digging} &    0.090    &   0.545    &   3.942    &   0.137    &   0.914   &    0.983    &   0.995\\
                Patil et al.~\cite{patil2020don}   &    0.087   &    0.495   &    3.775   &    0.133   &    0.917   &    0.983   &    0.995\\
                Packnet-SFM~\cite{guizilini20203d} &    0.078   &    0.420   &    3.485    &   0.121   &    0.931    &   0.986   &    0.996\\
                Johnston et al.~\cite{johnston2020self} & 0.081    &    0.484    &    3.716    &    0.126    &    0.927    &    0.985    &    0.996\\
                Wang et al.~\cite{wang2020self}&    0.082    &   0.462   &    3.739    &   0.127    &   0.923   &    0.984   &    0.996\\
				ManyDepth~\cite{watson2021temporal}    &   0.070  &   0.399  &   3.456  &   0.113  &   0.941  &   0.989  &   \underline{0.997}  \\
				Dynamicdepth~\cite{feng2022disentangling} &  \underline{0.068}    &     \underline{0.362}    &     3.454    &     \underline{0.111}    &     \underline{0.943}    &     \bf0.991    &     \bf0.998\\
                RA-depth~\cite{he2022ra} &    0.074    &     \underline{0.362}    &     \underline{3.345}    &     0.113     &    0.940    &     \underline{0.990}    &     \underline{0.997}\\
                \cline{1-8}
                \bf Ours     &   \bf0.064  &   \bf0.327  &   \bf3.170  &   \bf0.104  &   \bf0.950  &   \underline{0.990}  &   \underline{0.997}  \\
                \hline
			\end{tabular}
        }
\end{table}

\subsection{Full ablation study}
In Table~\ref{table:ablationapp}, we show the full ablation experiments on KITTI dataset. The experiment details are described in Sec.~\ref{sec:ablation}. Our final multi-frame depth network benefits from all the proposed methods. 
In Tables~\ref{table:speedapp} and ~\ref{table:dynamicapp}, we show the full metrics of the ablation experiments. The ablation details are described in Sec.~\ref{sec:ablation}. Our multi-scale feature fusion block improves the robustness of the depth estimation in larger camera ego-motions. Both the depth inconsistency mask and the robust knowledge distillation improves the depth estimation in dynamic regions.

\begin{table}[h]
    \small
    \renewcommand\arraystretch{1.1}
 	\caption{The full ablation results of on KITTI. Variations: (1) depth inconsistency mask, (2) multi-scale feature fusion, (3) consistent teacher, (4) robust mask.}
	\begin{center}
		\resizebox{0.475\textwidth}{!}
		{
			\begin{tabular}{|c|c|c|c|cccc|ccc|}
                \hline
				\multicolumn{4}{|c|}{Variations} & AbsRel & SqRel & RMSE & RMSE log & $\delta < 1.25$ & $\delta < 1.25^2$ &  $\delta < 1.25^3$\\
                \hline
				(1) &  (2) & (3) & (4) & \multicolumn{4}{c|}{$Lower\ is\ better$} & \multicolumn{3}{c|}{$Higher\ is\ better$}\\
				\hline
                \hline
				& & & & 0.098     &   0.770    &    4.459    &    0.176    &    0.900    &    0.965    &    0.983\\
                \cline{1-11} 
                  & $\surd$ & &   &  0.096  &   0.752  &   4.384  &   0.176  &   0.903  &   0.965  &   0.983  \\
				  $\surd$ & & &   & 0.096  &   0.735  &   4.369  &   0.173  &   0.902  &   0.966  &   0.984  \\
                 & & $\surd$ &   & 0.096  &   0.697  &   4.275  &   0.174  &   0.903  &   0.966  &   0.984  \\
                  $\surd$ &  $\surd$ & &   & 0.095  &   0.727  &   4.319  &   0.172  &   0.906  &   0.967  &   0.984  \\
                  & & $\surd$ & $\surd$  &   0.094  &   0.719  &   4.288  &   0.171  &   0.905  &   0.967  &   0.984  \\
                  $\surd$ &  & $\surd$ & $\surd$  & 0.093  &   0.700  &   4.254  &   0.170  &   0.908  &   0.967  &   0.984  \\
                  &  $\surd$ & $\surd$ & $\surd$ &  0.093  &   0.689  &   4.226  &   0.171  &   0.907  &   \bf0.968  &   0.984  \\
                \cline{1-11}
                  $\surd$ & $\surd$ & $\surd$  &$\surd$ & \bf0.090  &   \bf0.653  &   \bf4.173  &   \bf0.167  &   \bf0.911  &   \bf0.968  &   \bf0.985  \\ 
                \hline
			\end{tabular}
	   }
	\end{center}		
	\label{table:ablationapp}
\end{table}

\begin{table}[h]
	\centering
        \caption{The depth evaluation results on Cityscapes dataset with different test frames.  The longer time interval between adjacent frames implies larger camera ego-motions.}
    \renewcommand\arraystretch{1.1}
	\footnotesize
	\begin{center}
		\resizebox{0.475\textwidth}{!}
		{			\begin{tabular}{|c|c|cccc|ccc|}
                \hline
                Test & Multi-scale & AbsRel & SqRel & RMSE & RMSE log & $\delta < 1.25$ & $\delta < 1.25^2$ &  $\delta < 1.25^3$\\
				\cline{3-9}
                Frames &  Fusion & \multicolumn{4}{c|}{$Lower\ is\ better$} & \multicolumn{3}{c|}{$Higher\ is\ better$}\\
                \hline
               \hline
                & 
                &   \bf0.098  &   0.970  &   5.631 &   0.149  &   0.905  &   \bf0.977   &   \bf0.992  \\
                \multirow{-2}{*}{$I_t,I_{t-1}$}&  $\surd$  
                &   \bf0.098  &  \bf0.946  &   \bf5.553  &   \bf0.148  &   \bf0.908  &   \bf0.977  &   \bf0.992   \\
				\hline
                \hline
                & 
                &   0.101  &   0.998   &   5.670  &   0.152  &   0.903  &   0.976  &   0.992   \\
                \multirow{-2}{*}{$I_t,I_{t-2}$}&  $\surd$  
                &   \bf0.099  &   \bf0.960   &   \bf5.566  &   \bf0.149  &   \bf0.906  &   \bf0.977  &   \bf0.992  \\
				\hline
                \hline
                & 
                &   0.104  &   1.031   &   5.755  &   0.155  &   0.898  &   0.975  &   0.991   \\
                \multirow{-2}{*}{$I_t,I_{t-3}$}& $\surd$ 
                &   \bf0.101  &   \bf0.983   &   \bf5.625  &   \bf0.152  &   \bf0.903  &   \bf0.976   &   \bf0.992  \\
                \hline
			\end{tabular}
	   }
	\end{center}
	\label{table:speedapp}
\end{table}

\begin{table}[h]
	\centering
        \caption{The depth evaluation results of the dynamic objects (e.g. vehicles, bikes, and pedestrians) on the Cityscapes.}
    \renewcommand\arraystretch{1.1}
	\footnotesize
	\begin{center}
		\resizebox{0.475\textwidth}{!}
		{
			\begin{tabular}{|l|cccc|ccc|}
                \hline
                \multirow{2}{*}{Methods}  & AbsRel & SqRel & RMSE & RMSE log & $\delta < 1.25$ & $\delta < 1.25^2$ &  $\delta < 1.25^3$\\
				\cline{2-8}
                  & \multicolumn{4}{c|}{$Lower\ is\ better$} & \multicolumn{3}{c|}{$Higher\ is\ better$}\\
                \hline
               \hline
                ManyDepth~\cite{watson2021temporal} &   0.172   &   2.183  &   5.829  &   0.214  &   0.792  &   0.917  &   0.967  \\
                DynamicDepth~\cite{feng2022disentangling} &   0.145  &   1.547    &   5.143  &   0.183  &   0.830  &   0.950  &   0.983  \\
				\hline
                \hline
                Baseline &   0.178   &   2.469 & 6.123  &   0.222  &   0.787  &   0.911  &   0.962  \\
                Baseline + $M_i$ &   0.143   &   1.709 &   5.151  &   0.185  &   0.847  &   0.951   &   0.980  \\
                Baseline + $M_i$ + $M_r$ &  \bf 0.127  &  \bf 1.261  &  \bf 4.717  &  \bf 0.169  &  \bf 0.862  &  \bf 0.961 &   \bf 0.986  \\
                \hline
			\end{tabular}
	   }
	\end{center}
	\label{table:dynamicapp}
\end{table}

\section{Visualization Examples}

\subsection{Qualitative depth comparisons}
\begin{figure*}[!t]
    \centering
    \includegraphics[width=0.9\linewidth]{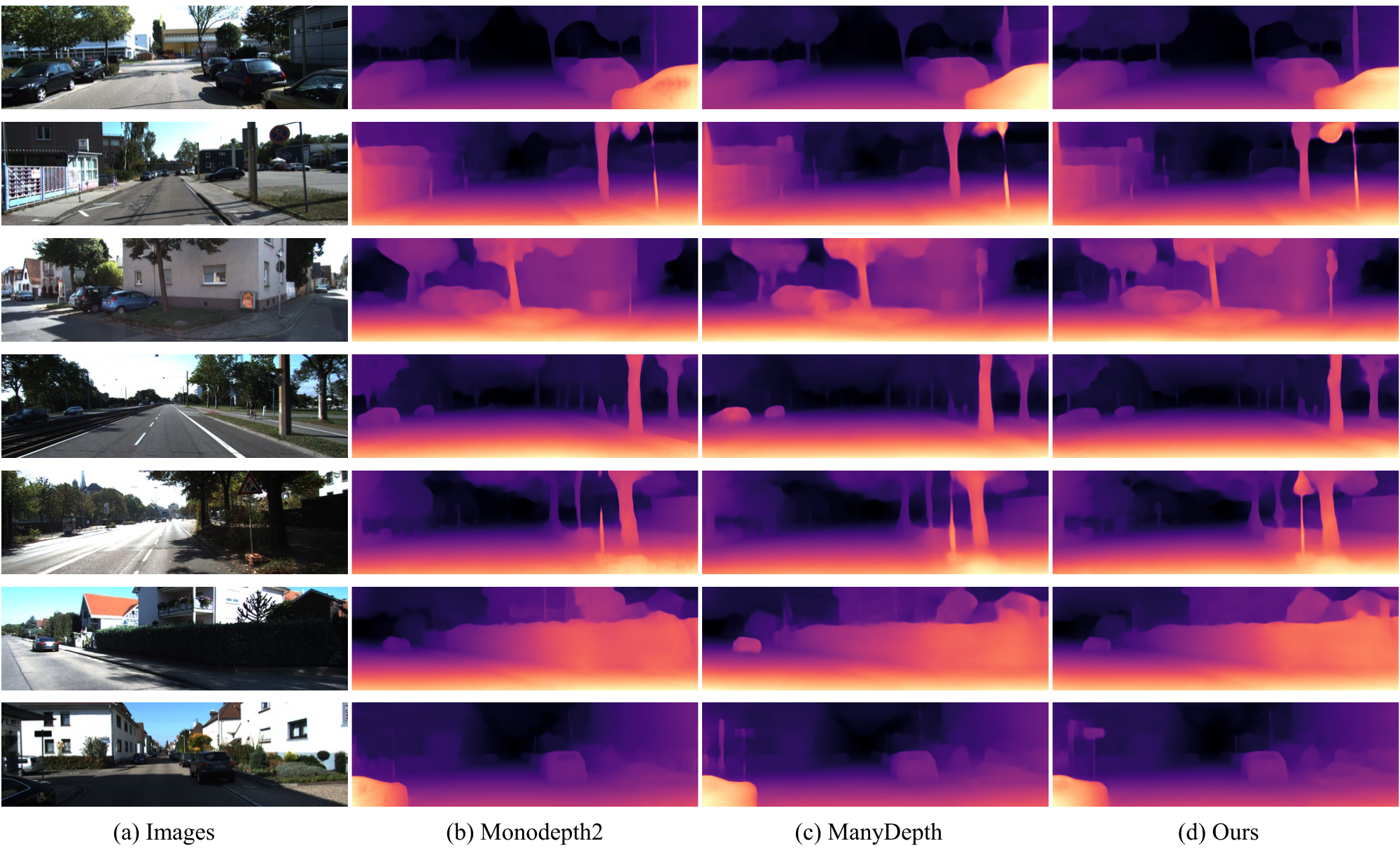}
    \vspace{-8pt}
    \captionof{figure}{Qualitative depth estimation comparison on KITTI. Our method performs better on dynamic objects and small-scale objects.}
    \label{fig:visual_all}

\end{figure*}

In Figure~\ref{fig:visual_all}, we show the qualitative comparison with previous methods. Our depth estimation performs better on dynamic objects and small-scale objects. Our depth estimation has sharper edges compared to previous methods.

\subsection{Qualitative depth inconsistency masks comparisons}
\begin{figure*}[!b]
    \centering
    {
    \includegraphics[width=0.9\linewidth]{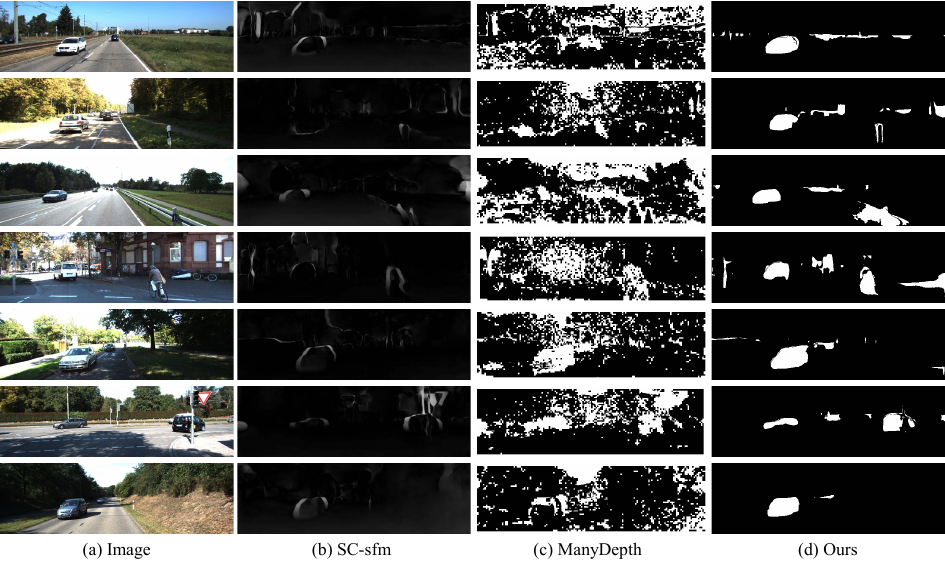}
    }
    \vspace{-8pt}
    \captionof{figure}{Qualitative depth inconsistency masks comparison.}
    \label{fig:maskcmp}

\end{figure*}

In Figure~\ref{fig:maskcmp}, we show the qualitative comparison with SC-sfm~\cite{bian2019unsupervised} and ManyDepth~\cite{watson2021temporal}. The inconsistency masks in SC-sfm~\cite{bian2019unsupervised} are based on the depth inconsistency between depth current frame and back-warped depth from adjacent frame. When there is occlusion, there is large depth inconsistency caused by difference between the depth of foreground objects and the depth of background. The dynamic objects will lead to larger occlusion regions. Therefore the sc-mask may detect some regions of the dynamic regions depending on the magnitude of the movement of the objects. The inconsistency masks in ManyDepth~\cite{watson2021temporal} are based on the depth inconsistency between single frame depth and min-cost depth. The minimal cost volume indicates that the corresponding depth may lead to best match. Due to the assumption of static scenes, there is difference in dynamic between the accurate depth and min-cost depth. Compared to these two methods, our method can recognize the dynamic regions more accurately and preserves more static regions.


\end{document}